%% file: main.tex
\documentclass[10pt,twocolumn,letterpaper]{article}
\usepackage{graphicx}
\usepackage{subcaption}
\usepackage[svgnames]{xcolor}
\usepackage{tikz}
\usepackage{graphicx}
\usepackage{subcaption}
\usepackage[svgnames]{xcolor}
\usepackage{tikz}
\usepackage{booktabs}
\usepackage[margin=1in]{geometry}
\usetikzlibrary{arrows.meta,positioning,fit,shadows.blur}
\usepackage{graphicx}
\usepackage{xcolor}
\usepackage{booktabs}
\usepackage{caption}

\usepackage{dblfloatfix}

\usepackage[pagenumbers]{cvpr}

\input{preamble}
\definecolor{cvprblue}{rgb}{0.21,0.49,0.74}
\usepackage[pagebackref,breaklinks,colorlinks,allcolors=cvprblue]{hyperref}

\def\confName{CVPR}
\def\confYear{2027}
\newcommand{\methodName}{WorldLatent\xspace}

\title{\methodName{}: A Unified 4D Latent Space for World Perception and Generation}

\author{
Jingxi Chen$^{1,2*}$ \quad
Yujia Chen$^{2}$ \quad
Cornelia Ferm\"{u}ller$^{1}$ \\
Yiannis Aloimonos$^{1}$ \quad
Wen-Sheng Chu$^{2\dagger}$ \quad
Du Tran$^{2\dagger}$ \\[0.6em]
$^{1}$University of Maryland, College Park \qquad
$^{2}$Google \\[2pt]
\url{http://WorldLatent.github.io}
}

\begin{document}
\twocolumn[{%
\renewcommand\twocolumn[1][]{#1}%
\maketitle
\begin{center}
    \centering
    \captionsetup{type=figure}
    \includegraphics[width=\textwidth]{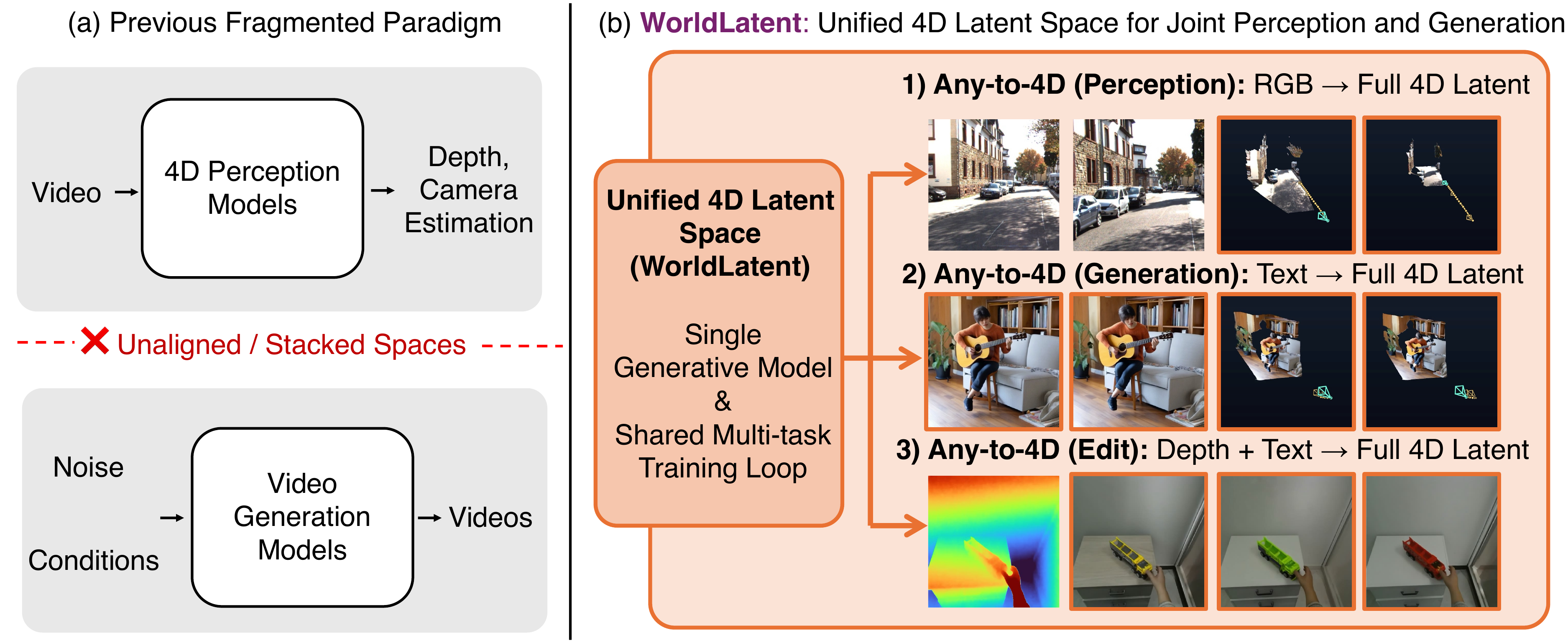}
    \caption{\textbf{Paradigm shift toward unified 4D world modeling.} (a) Prior paradigm: 4D perception and generation rely on separate one-way pipelines or unaligned, stacked modality spaces. (b) \methodName{} (Ours): Our unified 4D latent space enables a single generative model and shared multi-task training loop to jointly execute 1) 4D perception (RGB $\rightarrow$ 3D geometry and camera poses), 2) Text-to-4D generation (text $\rightarrow$ RGB video, 3D geometry, and camera motion), and 3) Controllable 4D editing (depth + text $\rightarrow$ diverse appearance-edited videos, e.g., recolored trucks). \textit{Orange borders} denote target task outputs decoded from the generated full 4D latent.}
\label{fig:teaser}
\end{center}%
 }]
{\renewcommand{\thefootnote}{*}\footnotetext{This work was done during Jingxi Chen's internship at Google.}}
{\renewcommand{\thefootnote}{$\dagger$}\footnotetext{Joint last authors.}}
\addtocontents{toc}{\protect\setcounter{tocdepth}{-1}}
\input{sec/0_abstract}

\input{sec/1_intro}

\input{sec/2_related_work}

\input{sec/3_method}
\input{sec/4_experiments}
\input{sec/5_conclusion}

\newpage

{
    \small
    \bibliographystyle{ieeenat_fullname}
    \bibliography{main}
}

\maketitlesupplementary
\appendix
\addtocontents{toc}{\protect\setcounter{tocdepth}{3}}
\tableofcontents

\section{Video Results}
Please refer to our project webpage for video results, which clearly demonstrate the effectiveness of our method and its comparison with baseline models.

\section{More Training and Dataset Details}
\subsection{Dataset Details}
\noindent\textbf{Multi-Domain Training Mixture.}
To train both \textit{WorldVAE} (Stage~1) and our unified \textit{Any-to-4D} diffusion model (Stage~2), we curate approximately $1.4$M $17$-frame video clips spanning $22$ diverse data sources, summarized in Table~\ref{tab:supp_datasets}. Our sampling distribution balances two complementary pools:
\begin{itemize}
    \item \textbf{Real-World Video Pool ($79.4\%$ of samples):} Draws from large-scale open-domain and text-video corpora (SpatialVid-HQ~\cite{wang2026spatialvid}, VidGen-1M~\cite{tan2024vidgen}, Kinetics-600~\cite{kay2017kinetics}, OpenVid-1M~\cite{nan2025openvid}, Open-Sora~\cite{zheng2024open}), camera- and object-centric real videos (RealCam-Vid~\cite{zheng2025realcam}, SA-V~\cite{ravi2025sam}, CO3D~\cite{reizenstein2021common}), and egocentric or robotic manipulation streams (Ego-Exo4D~\cite{grauman2024ego}, EgoDex~\cite{hoque2026egodex}, HOI4D~\cite{liu2022hoi4d}, DROID~\cite{khazatsky2024droid}). This pool provides broad open-world visual concepts, complex non-rigid motion, realistic lighting, and diverse scene coverage essential for generalizable 4D generation and editing, with dense depth and camera pose pseudo-labels generated by DA3-Giant~\cite{lin2025depth}.
    \item \textbf{Labeled 3D/4D Pool ($20.6\%$ of samples):} Combines synthetic 4D datasets with exact rendering-engine depth and camera trajectories (PointOdyssey~\cite{zheng2023pointodyssey}, OmniWorld~\cite{zhou2025omniworld}, Hypersim~\cite{roberts2021hypersim}, TartanAir~\cite{wang2020tartanair}, Dynamic Replica~\cite{karaev2023dynamicstereo}, VKITTI2~\cite{cabon2020virtual}, MVS-Synth~\cite{huang2018deepmvs}) alongside calibrated real-sensor 3D scans (WildRGB-D~\cite{xia2024rgbd}, ScanNet++~\cite{yeshwanth2023scannet++}, Waymo Open~\cite{sun2020scalability}). These datasets anchor sharp geometric boundaries, accurate occlusion disambiguation, and physically grounded 3D camera motion. Because real-sensor 3D scans often contain sensor noise and incomplete depth regions, we likewise use DA3-Giant~\cite{lin2025depth} as a teacher to generate clean, dense 4D labels for them.
\end{itemize}
Across both pools, we use Gemini 3.5 Flash \cite{team2023gemini} to generate uniform, high-quality text captions for all video clips. Crucially, all zero-shot perception benchmarks evaluated in the main paper, MPI Sintel~\cite{butler2012naturalistic}, Bonn RGB-D Dynamic~\cite{palazzolo2019refusion}, and ScanNet~\cite{dai2017scannet}, as well as the $55$ held-out clips used for generation evaluation, are strictly excluded from all Stage~1 and Stage~2 training splits.

\subsection{Training Details}
\noindent\textbf{Model Parameters.}
For WorldVAE ($\sim$528M parameters), we initialize the four modality encoders and decoders from the Wan2.1-T2V VAE ($4 \times 127$M), keeping the encoders frozen while fully fine-tuning the decoders alongside a 20M-parameter DPT Fusion module. For Any-to-4D, we perform full-parameter fine-tuning of Wan2.1-T2V-14B (14B) for our final model and use Wan2.1-T2V-1.3B (1.3B) for controlled ablations. All models generate 17-frame 4D sequences at $512 \times 512$ resolution across RGB, depth, and Pl\"ucker ray maps. 

\noindent\textbf{Training Details.}
We train our final 14B model on $256$ TPU accelerators ($32$~GB memory each) for $108$k steps with a global batch size of $256$. Optimization uses AdamW~\cite{loshchilov2017decoupled} ($\beta_2 = 0.95$, weight decay $0$) with a learning rate of $3 \times 10^{-5}$. During joint multi-task training, we sample tasks at each iteration with probabilities of $0.45$ for text-to-4D generation, $0.30$ for depth-conditioned 4D editing/re-rendering, and $0.25$ for monocular 4D perception. During inference, we use Euler sampler \cite{esser2024scaling, liu2022flow, lipman2022flow} with flow shift 3. For Text-to-4D: we do 25 diffusion steps with Classifier Free Guidance (CFG) 4, CFG-rescale 0.7. For Render: we do 25 steps, with CFG 2. For Perception: we do 8 steps, with empty text, no guidance. For WorldVAE, we train with a learning rate of $5 \times 10^{-5}$. We set the reconstruction weights to $\lambda_1 = 4$ (RGB), $\lambda_2 = 2$ (depth), and $\lambda_3 = \lambda_4 = 1$ (Pl\"ucker ray maps), alongside $\lambda_{\text{p}} = 1$ (LPIPS), $\lambda_{\text{g}} = 1$ (depth gradient), $\lambda_{\text{pts}} = 0.5$ (3D pointmap), and $\lambda_{\text{KL}} = 1 \times 10^{-5}$. To support partial-modality conditioning in Stage~2, each training sample is encoded from a randomly drawn modality subset (replacing omitted streams with a learned mask token) with probabilities for inputs given to the encoder: $0.40$ for all modalities (RGB + depth + rays), $0.20$ for RGB only, $0.15$ for depth only, $0.15$ for rays only, $0.05$ for RGB + depth, and $0.05$ for RGB + rays.

\smallskip
\noindent\textbf{DPT Fusion Architecture ($\mathcal{F}_{\text{DPT}}$).}
Our $20.7$M-parameter DPT Fusion module adapts DPT~\cite{ranftl2021vision} to spatiotemporal latent fusion from scratch. Four $16$-ch Wan2.1-VAE~\cite{wan2025wan} latents ($5{\times}64{\times}64$, patch size $1$) receive learned modality embeddings and shared positional embeddings before entering a $12$-layer scratch ViT ($d{=}192$, $6$ heads, $8.6$M), where missing modalities use a learned mask token. At four tap depths, the four modality tokens per cell are concatenated and MLP-fused, processed by zero-initialized temporal mixers ($2$ temporal blocks/tap + $2$ global cross-frame blocks, $7.2$M), and aggregated via a non-upsampling DPT pyramid (scales $1\times, 0.5\times, 0.25\times, 0.25\times$, widths $48/96/192/384$, $128$-ch neck, $4.9$M) into $64$ channels (\text{mean} $\boldsymbol{\mu}_{\text{4d}}, \text{and log-variance} \log\boldsymbol{\sigma}_{\text{4d}}^2$) for the $32$-ch $\mathbf{z}_{\text{4d}}$. For decoding, a modality-conditioned conv and per-modality $1{\times}1$ projections ($0.03$M) map $\mathbf{z}_{\text{4d}}$ back to $16$-ch streams for decoders $D_1$--$D_4$.

\begin{table}[t]
\centering
\footnotesize
\setlength{\tabcolsep}{0pt}
\renewcommand{\arraystretch}{1.10}
\begin{tabular*}{\columnwidth}{@{\extracolsep{\fill}}lr@{}}
\toprule
\textbf{Dataset} & \textbf{\% Total} \\
\midrule
\multicolumn{2}{@{}l}{\textit{Real-World Video Pool ($79.4\%$ of samples)}} \\
\midrule
SpatialVid-HQ~\cite{wang2026spatialvid} & $\sim\!19.1$ \\
VidGen-1M~\cite{tan2024vidgen} & $\sim\!15.1$ \\
Kinetics-600~\cite{kay2017kinetics} & $\sim\!11.9$ \\
OpenVid-1M~\cite{nan2025openvid} & $\sim\!10.7$ \\
RealCam-Vid~\cite{zheng2025realcam} & $\sim\!8.7$ \\
SA-V~\cite{ravi2025sam} & $\sim\!4.7$ \\
Open-Sora~\cite{zheng2024open}, CO3D~\cite{reizenstein2021common} & $\sim\!1.6$ each \\
Ego-Exo4D~\cite{grauman2024ego}, EgoDex~\cite{hoque2026egodex} & $\sim\!1.5$ each \\
HOI4D~\cite{liu2022hoi4d}, DROID~\cite{khazatsky2024droid} & $\sim\!1.5$ each \\
\cmidrule{1-2}
\textbf{Real-World Video Subtotal} & $\mathbf{79.4\%}$ \\
\midrule
\multicolumn{2}{@{}l}{\textit{Labeled 3D/4D Pool ($20.6\%$ of samples)}} \\
\midrule
WildRGB-D~\cite{xia2024rgbd} (DA3) & $4.8$ \\
PointOdyssey~\cite{zheng2023pointodyssey} & $4.8$ \\
ScanNet++~\cite{yeshwanth2023scannet++} (DA3) & $4.7$ \\
OmniWorld~\cite{zhou2025omniworld} & $1.7$ \\
Hypersim~\cite{roberts2021hypersim} & $1.4$ \\
TartanAir~\cite{wang2020tartanair} & $1.0$ \\
Dynamic Replica~\cite{karaev2023dynamicstereo} & $0.9$ \\
VKITTI2~\cite{cabon2020virtual} & $0.5$ \\
Waymo Open~\cite{sun2020scalability} (DA3) & $0.5$ \\
MVS-Synth~\cite{huang2018deepmvs} & $0.3$ \\
\cmidrule{1-2}
\textbf{Labeled 3D/4D Subtotal} & $\mathbf{20.6\%}$ \\
\midrule
\textbf{Total ($\sim$1.4M 17-frame clips)} & $\mathbf{100.0\%}$ \\
\bottomrule
\end{tabular*}
\caption{\textbf{Training data mixture ($\sim$1.4M clips, \% of all samples).} Real-world videos ($79.4\%$) and \texttt{(DA3)}-marked 3D datasets use DA3-Giant~\cite{lin2025depth} depth and camera pose pseudo-labels.}
\label{tab:supp_datasets}
\end{table}

\section{Additional Visual Results}
\subsection{Additional 4D Perception Visual Results}

\begin{figure*}[t]
    \centering
    \includegraphics[width=\textwidth,height=0.9\textheight,keepaspectratio]{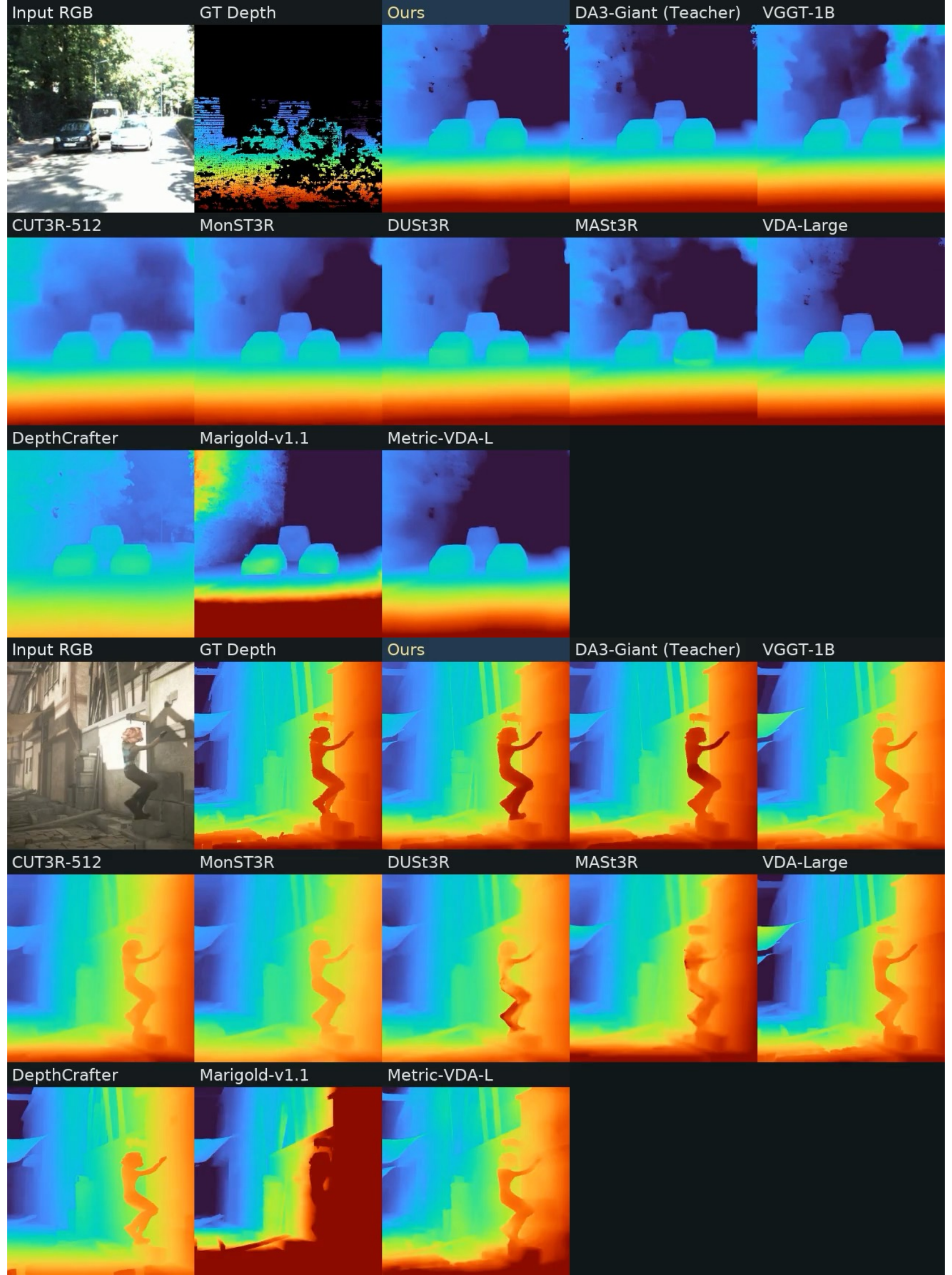}
    \caption{\textbf{Additional 4D perception visual results.} Qualitative video depth comparison on KITTI~\cite{geiger2012we} (Top) and Sintel (Bottom). Panels (from left to right, top to bottom) show: input RGB, ground-truth depth, \methodName{}-14B (Ours), DA3-Giant (Teacher), VGGT-1B, CUT3R-512, MonST3R, DUSt3R, MASt3R, Video-Depth-Anything-L, DepthCrafter, Marigold-v1.1, and Metric-VDA-Large.}
\label{fig:supp_percep_1}

\end{figure*}

\begin{figure*}[t]
    \centering
    \includegraphics[width=\textwidth,height=0.9\textheight,keepaspectratio]{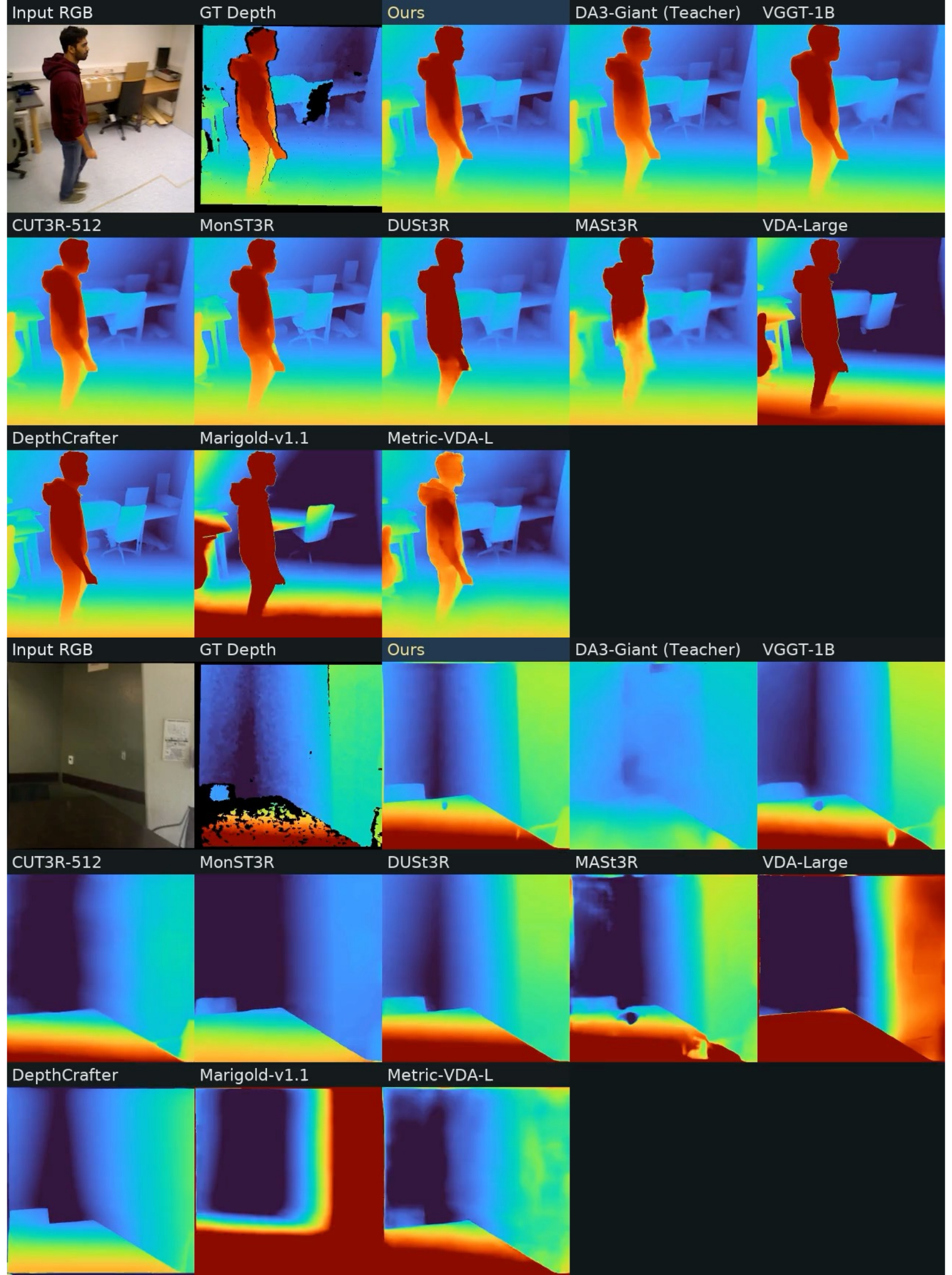}
    \caption{\textbf{Additional 4D perception visual results.} Qualitative video depth comparison on Bonn (Top) and ScanNet (Bottom), panel layout follows Fig.~\ref{fig:supp_percep_1}.}
    \label{fig:supp_percep_2}
\end{figure*}

Figures~\ref{fig:supp_percep_1} and~\ref{fig:supp_percep_2} provide additional qualitative comparisons of our depth perception results against the baselines. Please refer to the accompanying website page for full video results.

\subsection{Additional Re-rendering Results}
\label{sec:supp_render}

\begin{figure*}[t]
    \centering
    \includegraphics[width=\textwidth,height=0.9\textheight,keepaspectratio]{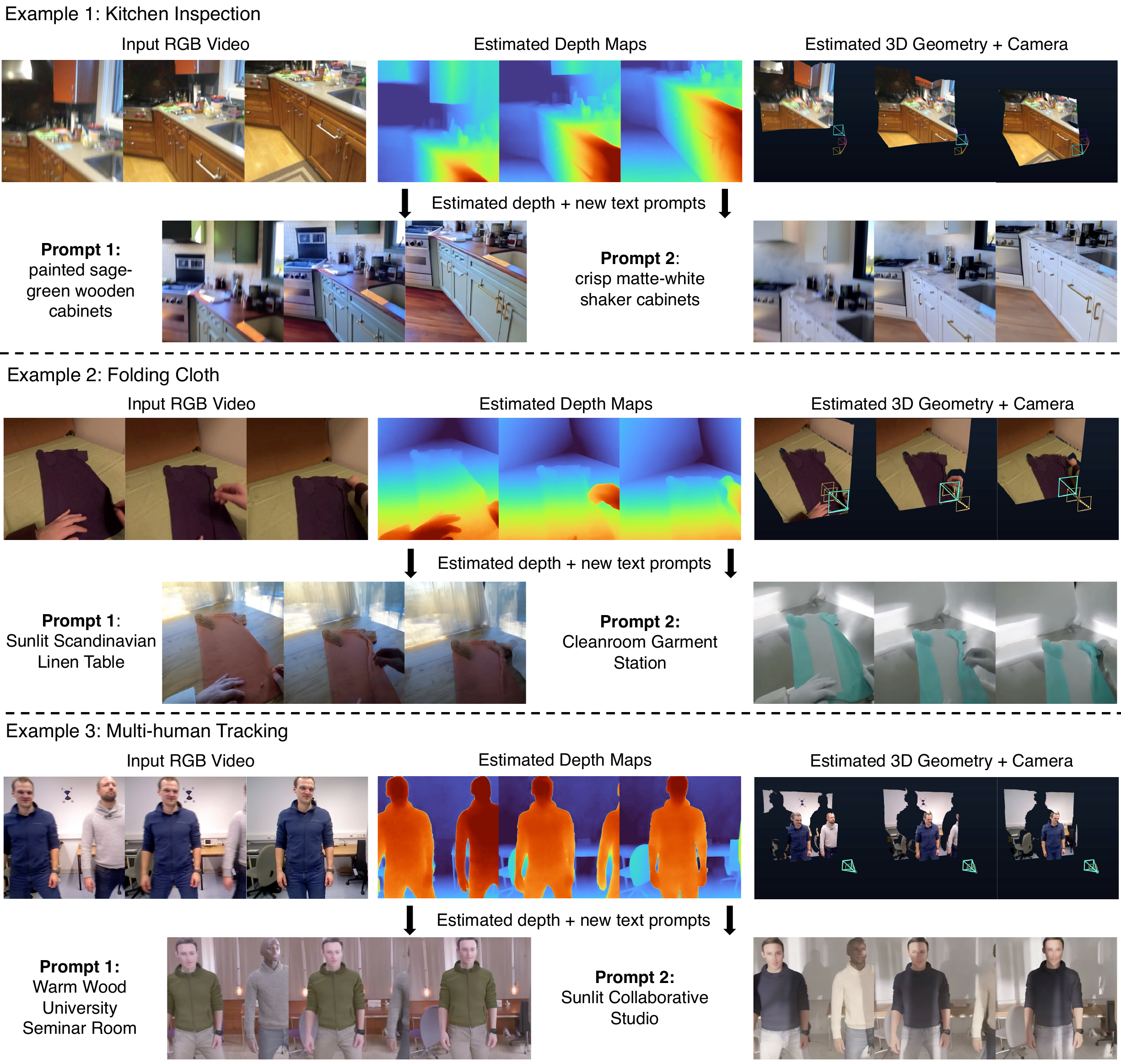}
    \caption{\textbf{Additional depth-conditioned re-rendering results.} For each input RGB video (kitchen inspection, cloth folding, and multi-human tracking), we show our estimated depth maps and 3D geometry with camera poses (top row). The estimated depth is then combined with two new text prompts to re-render the scene (bottom row), changing its appearance while keeping its geometry and motion.}
    \label{fig:supp_render}
\end{figure*}

Figure~\ref{fig:supp_render} presents additional depth-conditioned 4D re-rendering results across three diverse scenes. Using a single \methodName{}-14B checkpoint, we first perform monocular 4D perception on each input RGB video to estimate dense depth and camera poses, and then condition on the predicted depth alongside two distinct text prompts to synthesize geometrically consistent, appearance-edited videos. Please refer to the accompanying website page for full video results.

\subsection{Additional Text-based 4D Generation Results}
\label{sec:supp_gen}
\begin{figure*}[t]
    \centering
    \includegraphics[width=\textwidth,height=0.9\textheight,keepaspectratio]{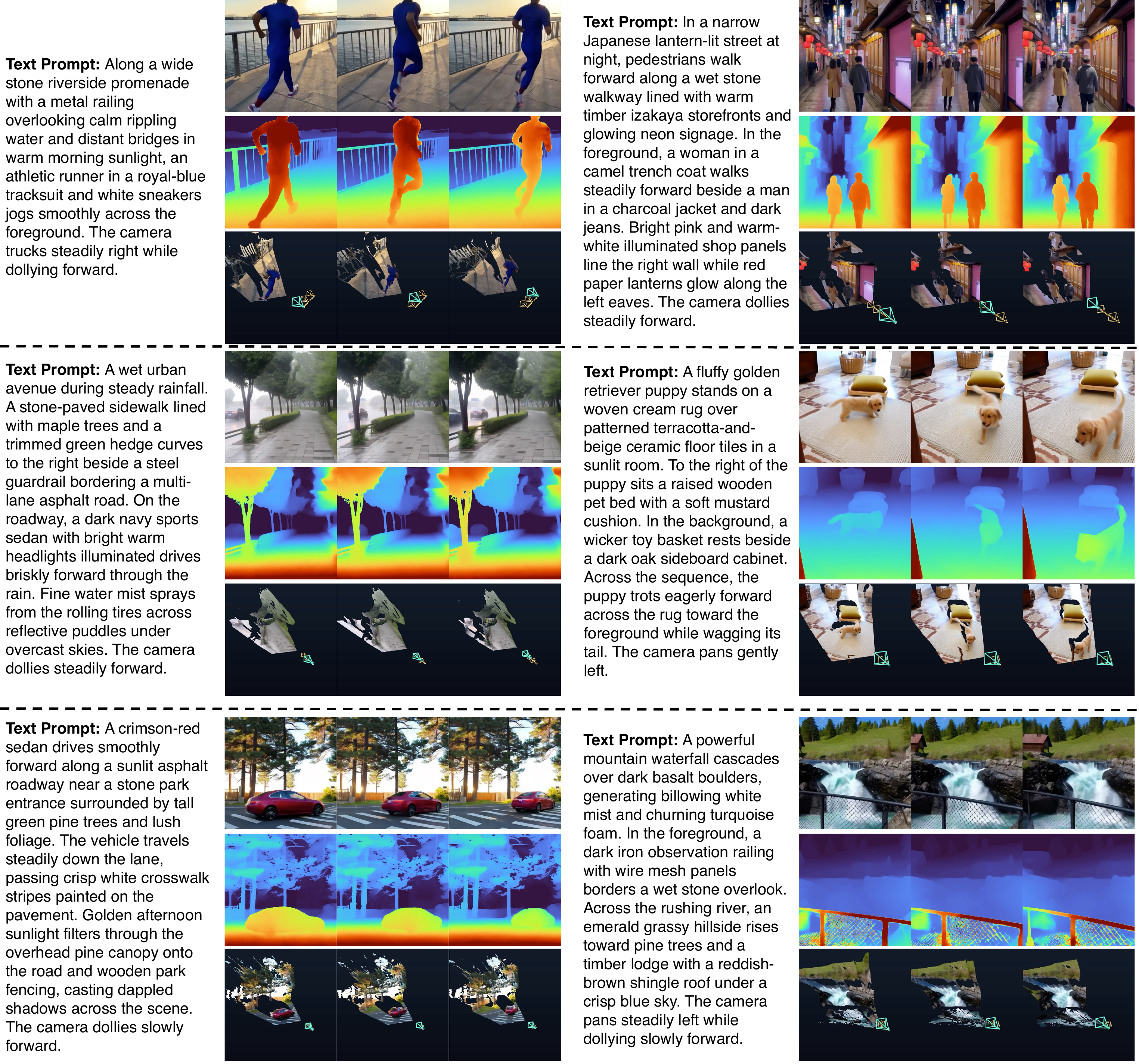}
    \caption{\textbf{Additional text-based 4D generation results.} For each text prompt, we show the generated RGB video (top), depth maps (middle), and 3D geometry with camera poses (bottom).}
    \label{fig:supp_gen}
\end{figure*}

Figure~\ref{fig:supp_gen} shows additional text-based 4D generation results. Given only a text prompt, our model generates the RGB video together with its depth maps and camera trajectory, from which the 3D geometry is recovered. Please refer to the accompanying website page for full video results.

\end{document}

%% file: sec/0_abstract.tex
\begin{abstract}
Generative Pre-Training (GPT) represents a profound milestone in natural language processing, serving as the foundation for most of today's state-of-the-art large language models. By casting diverse understanding and generation problems into a shared token representation, GPT enables a single architecture and unified optimization loop to master heterogeneous tasks across massive corpora, giving rise to powerful foundation models for natural language. In contrast, 4D spatiotemporal vision lacks a common representational protocol: perception (such as depth and camera estimation) and generation are either handled by separate one-way pipelines or bolted together from unaligned single-modality spaces. In this work, we introduce WorldLatent, a framework that formulates a unified 4D latent space as a common protocol for both world perception and generation. Rather than relying on fragmented, task-specific models or stacked single-modality representations, our framework proposes a multimodal VAE architecture that compresses and fuses complex 4D visual appearance, geometry, and dynamics into a single structured latent representation. Within this formulation, perception and generation operate as complementary denoising processes over the same latent space, enabling one generative model to learn both capabilities simultaneously within a single training loop. Experiments across representative 4D benchmarks verify that our fused latent protocol achieves specialist-level zero-shot 4D perception, coherent 4D generation, and controllable editing from a single model. By demonstrating that diverse 4D tasks can be seamlessly unified under a shared representational protocol, our work provides a scalable foundation for general-purpose 4D spatial intelligence.

\end{abstract}

%% file: sec/1_intro.tex
\section{Introduction}
\label{sec:intro}

The establishment of a unified task protocol in Generative Pre-Training (GPT)~\cite{radford2019language, raffel2020exploring, brown2020language} marked a transformative milestone in natural language processing. Rather than relying on specialized architectures and loss functions~\cite{kim2014convolutional, sutskever2014sequence, seo2016bidirectional, devlin2019bert} for text classification, machine translation, question answering, and open-ended synthesis, casting these diverse problems into a shared token representation enabled a single model to learn both understanding and generation within one optimization objective. Inspired by this success, we argue that achieving general-purpose 4D spatial intelligence in computer vision requires a similar unification: an intelligent visual system must not only perceive the 3D geometry and camera motion of a dynamic scene from video, but also imagine, synthesize, and edit coherent 4D worlds over time. Yet, in contrast to language modeling, 4D spatiotemporal vision still lacks such a common representational protocol.

As illustrated in Figure~\ref{fig:teaser}(a), contemporary 4D perception and video generation still rely on separate one-way pipelines or unaligned, stacked modality spaces. On the perception side, feed-forward geometric reconstructors~\cite{wang2024dust3r, leroy2024grounding, zhang2025monst3r, wang2025continuous, wang2025vggt, lin2025depth} and diffusion-based depth or pose estimators~\cite{ke2024repurposing, hu2025depthcrafter, xu2025geometrycrafter, zhang2024cameras}, including recent adaptations of video generators for geometry estimation~\cite{jiang2025geo4d, sun2025unigeo}, map videos to depth, point clouds, or camera trajectories, yet overwrite or omit RGB generation and cannot synthesize new visual worlds. On the generation side, large-scale video Diffusion Transformers (DiTs)~\cite{brooks2024video, yang2025cogvideox, wan2025wan} synthesize compelling imagery in 2D RGB latent spaces without explicit 3D grounding, geometry-guided generators~\cite{ren2025gen3c, wu2026geometry} inject external 3D caches or feature alignment unidirectionally while still decoding solely into 2D pixels. Even recent frameworks that jointly diffuse appearance and geometry within one backbone~\cite{zhang2025world, zhu2025aether, kiruga2026ugg} bypass multimodal tokenizer learning, heuristically passing geometric maps through frozen RGB Variational Autoencoders (VAEs) and concatenating disjoint latents. Bridging these two paradigms properly is non-trivial: geometric estimation methods seek representations that capture high-level structural abstraction to infer scene geometry and camera motions, whereas video generation models rely on generative compression that preserves low-level photometric details for high-fidelity visual reconstruction. This tension between geometric abstraction and generative compression motivates learning a genuinely fused representation rather than stacking separate latents.

In this work, we take a step toward scaling general-purpose 4D spatial intelligence by tackling the fundamental question: \textit{``How can diverse knowledge of 4D world perception and generation be consolidated into a single generative model with a shared multi-task training loop?''} Inspired by task unification in large language models, we seek a shared 4D representational protocol across visual appearance, geometry, and dynamics to bridge these capabilities.

To realize this protocol, we introduce \textit{\methodName{}} (Figure~\ref{fig:teaser}(b)), a framework that establishes a unified 4D latent space as a shared spatiotemporal representation for world perception and generation. Rather than relying on one-way pipelines or stacking unaligned modality latents, \methodName{} parameterizes a dynamic scene jointly via RGB video, dense depth, and pixel-aligned camera ray maps. To reconcile geometric structure with high-fidelity appearance compression, we design a multimodal 4D Variational Autoencoder (VAE) that compresses and fuses these modalities into a single spatiotemporal latent grid at the same token resolution as standard video VAEs.

Built upon this shared representation, \methodName{} unifies perception, generation, and controllable editing as complementary \textit{Any-to-4D} denoising processes within a single generative model: (1) Any-to-4D Perception (RGB
$\rightarrow$ Full 4D Latent), which conditions on an uncalibrated monocular video and denoises the full 4D latent to recover depth, dynamic 3D point clouds, and camera trajectories, (2) Any-to-4D Generation (Text 
$\rightarrow$ Full 4D Latent), which jointly denoises appearance, depth, and camera rays from noise to synthesize coherent videos paired with aligned 3D scene geometry and camera motion, and (3) Any-to-4D Editing and Rendering (Depth + Text 
$\rightarrow$ Full 4D Latent), which conditions on 4D depth structure and text prompts to render or edit visual appearance with strict 3D geometric adherence. Because every capability operates over the same latents with identical input and output spaces, a generative backbone can be adapted into a general-purpose 4D world model within a single multi-task training loop across perception and generation.

We instantiate and validate \methodName{} on a 14B-parameter generative model trained jointly across 4D perception, generation, and controllable editing. Evaluating across indoor, outdoor, and dynamic 4D benchmarks (MPI Sintel~\cite{butler2012naturalistic}, Bonn~\cite{palazzolo2019refusion}, and ScanNet~\cite{dai2017scannet}) as well as text-to-4D and depth-guided synthesis, we show that our fused 4D latent representation enables joint multi-task denoising within a single model, delivering strong geometric estimation, faithful geometry-guided rendering, and coherent 4D generation over a shared input-and-output latent space.

In summary, our main contributions are threefold:
\begin{itemize}
    \item We propose \methodName{}, a unified 4D latent protocol for world perception and generation, employing a multimodal VAE to compress and fuse 4D visual appearance, geometry, and dynamics into a single shared latent space for diverse downstream 4D tasks.
    \item We formulate a unified \textit{Any-to-4D} generative modeling paradigm that consolidates 4D perception, generation, and editing into a single model with a shared multi-task training loop over the \methodName{} representation.
    \item Across controlled ablations and a 14B-parameter unified model, we demonstrate that a single generative model trained on \methodName{} effectively supports dynamic 4D perception, 4D scene generation, and controllable neural rendering, establishing a scalable foundation for general-purpose 4D spatial intelligence.
\end{itemize}

%% file: sec/2_related_work.tex
\begin{figure*}[t!]
    \centering
    \includegraphics[width=\textwidth]{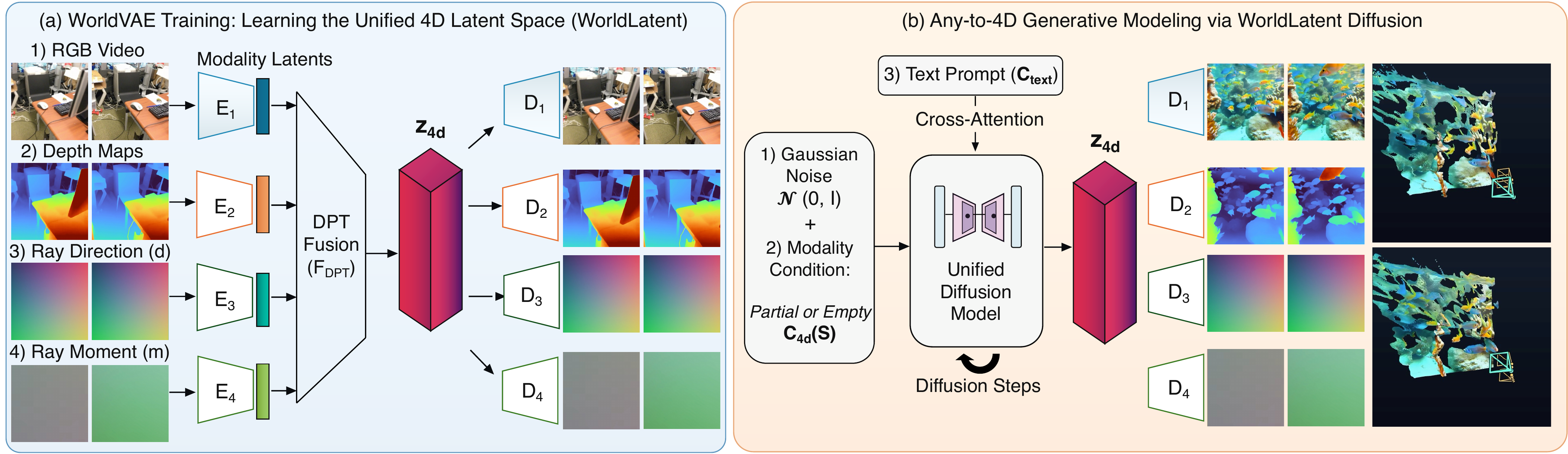}
    \caption{\textbf{\methodName{} training pipeline}. (a) WorldVAE training: Multiple 4D spatiotemporal modalities, including monocular RGB video, depth maps, and camera Pl\"ucker coordinates (ray direction $\mathbf{d}$ and moment $\mathbf{m}$), are compressed via modality-specific encoders ($E_1$--$E_4$) and fused by a Dense Prediction Transformer (DPT) into a unified latent bottleneck, $\mathbf{z}_{\text{4d}}$ (\methodName{}). Modality-specific decoders ($D_1$--$D_4$) then reconstruct the original modalities from this shared latent space. (b) Any-to-4D generative modeling: With the WorldVAE frozen, a unified diffusion model is trained within the learned latent space. It takes Gaussian noise, text prompts ($C_{\text{text}}$), and partial or empty 4D latents $\mathbf{c}_{\text{4d}}(\mathcal{S})$ (e.g., $\mathcal{S} = \varnothing$ for text-to-4D generation, RGB-only for perception, or depth-only for editing) to synthesize the full 4D latent $\mathbf{z}_{\text{4d}}$. Finally, the frozen decoders ($D_1$--$D_4$) map $\mathbf{z}_{\text{4d}}$ back into temporally coherent video, 3D geometry, and camera trajectories.}
    \label{fig:pipeline}
\end{figure*}

\section{Related Work}
\label{sec:related_work}

\noindent\textbf{3D and 4D Spatiotemporal Perception.}
Feed-forward pointmap regressors~\cite{wang2024dust3r, wang2025vggt, lin2025depth, leroy2024grounding} predict 3D point clouds without structure-from-motion, with extensions to dynamic and streaming videos~\cite{zhang2025monst3r, wang2025continuous}. Meanwhile, monocular~\cite{yang2024depth, chen2025video} and diffusion~\cite{ke2024repurposing, hu2025depthcrafter, shao2025learning} depth models recover 2.5D depth without camera poses, and video-diffusion adaptations~\cite{xu2025geometrycrafter, jiang2025geo4d, sun2025unigeo} estimate multi-frame 3D/4D geometry. 

\smallskip
\noindent\textbf{Video and 4D World Generation.}
Video DiTs~\cite{brooks2024video, yang2025cogvideox, wan2025wan} lack 3D grounding in RGB latents, causing static-camera bias or drift. Conditioned generators inject camera, depth, or 3D cache cues~\cite{he2025cameractrl, zhang2023adding, esser2023structure, ren2025gen3c}, or align features to geometric teachers~\cite{wu2026geometry}, unidirectionally without inferring 4D geometry, while dedicated 4D generators~\cite{wu2025cat4d, bahmani20244d, ling2024align, xie2025sv4d, fang2025worldreel} remain synthesis specialists. Although WVD~\cite{zhang2025world} and Aether~\cite{zhu2025aether} jointly diffuse appearance and geometry, WVD targets image-conditioned static scenes and Aether drops text conditioning on synthetic-only data, neither supports general-purpose multi-task learning across open-world 4D perception, generation, and editing under a unified representation protocol.

\smallskip
\noindent\textbf{Unified Multi-Task Representation Protocols.}
Inspired by unified task protocols in NLP~\cite{raffel2020exploring} and multimodal vision~\cite{lu2022unified, mizrahi20234m, bai2024sequential}, recent works explore joint latent spaces: Orchid~\cite{krishnan2025orchid} encodes RGB, depth, and normals for static 2D images, requiring separate image-conditioned fine-tuning for perception, while MotionCrafter~\cite{zhu2026motioncrafter} learns a 4D geometry-and-motion VAE without visual synthesis. In 4D world modeling, joint frameworks~\cite{zhang2025world, zhu2025aether, kiruga2026ugg} heuristically stack frozen RGB-VAE and geometric channels, retaining cross-modal redundancy and rigid modality boundaries. Instead, WorldVAE fuses RGB video, depth, and camera rays into a single compact bottleneck at an RGB-only VAE's token count, enabling the unification of 4D perception, generation, and editing within one model and multi-task training.

%% file: sec/3_method.tex
\section{Method}
\label{sec:method}

Our goal is to establish a unified representational protocol that bridges 4D physical perception and generative modeling within a single model and a single optimization loop. As illustrated in Figure \ref{fig:pipeline}, our \methodName{} training pipeline consists of two stages: (a) Unified 4D Latent Space via multimodal WorldVAE training and (b) Unified Any-to-4D Generative Modeling via \methodName{} diffusion training.

\subsection{\methodName{}: Unified 4D Latent Space}
\label{sec:stage1}

\noindent\textbf{Spatiotemporal 4D Modality Representation.}
A dynamic 3D scene observed over $T$ frames at spatial resolution $H \times W$ comprises visual appearance, 3D scene geometry, and camera motion. 
To align all modalities onto a common spatiotemporal grid, we parameterize camera trajectories using dense Pl\"ucker ray coordinates \cite{zhang2024cameras}. For each pixel $(u, v)$ at frame $t$, its camera ray is uniquely defined by a unit ray direction $\mathbf{d}_{t,u,v} \in \mathbb{R}^3$ and a ray moment $\mathbf{m}_{t,u,v} = \mathbf{o}_t \times \mathbf{d}_{t,u,v} \in \mathbb{R}^3$, where $\mathbf{o}_t \in \mathbb{R}^3$ denotes the camera optical center in world coordinates. Thus, any 4D scene is represented by four spatiotemporally aligned tensors:
(1) \textit{Visual Appearance (RGB Video):} $\mathbf{X}_1 = \mathbf{I} \in \mathbb{R}^{T \times H \times W \times 3}$,
(2) \textit{Scene Geometry (4D Depth Map):} $\mathbf{X}_2 = \mathbf{D} \in \mathbb{R}^{T \times H \times W \times 1}$,
(3) \textit{Camera Ray Direction:} $\mathbf{X}_3 = \mathbf{d} \in \mathbb{R}^{T \times H \times W \times 3}$, and
(4) \textit{Camera Ray Moment:} $\mathbf{X}_4 = \mathbf{m} \in \mathbb{R}^{T \times H \times W \times 3}$.

\smallskip
\noindent\textbf{Modality-Specific Encoding and Fusion.}
As shown in Figure~\ref{fig:pipeline} (a), each modality $\mathbf{X}_i$ ($i \in \{1, 2, 3, 4\}$) is first processed by a dedicated spatiotemporal encoder $E_i$, yielding modality-specific latent features $\mathbf{f}_i = E_i(\mathbf{X}_i) \in \mathbb{R}^{T' \times H' \times W' \times C} $ downsampled by temporal and spatial factors ($f_t, f_s$) to the resolution $T' \times H' \times W'$ and the latent channel $C$. Rather than concatenating modality latents naively, which preserves cross-modal redundancy and inflates channel dimensionality, we introduce a Dense Prediction Transformer (DPT)~\cite{ranftl2021vision} for cross-modality fusion $\mathcal{F}_{\text{DPT}}$. Specifically, at four intermediate ViT tap depths, the four modality token streams at each spatiotemporal patch are concatenated channel-wise and projected into a single fused token grid via a two-layer GELU MLP~\cite{hendrycks2016gaussian} with LayerNorm~\cite{ba2016layer}. These multi-scale fused features are then progressively aggregated to parameterize the posterior distribution of our Unified 4D Latent Space (\methodName{}):
\begin{equation}
    \mathbf{z}_{\text{4d}} = \mathcal{F}_{\text{DPT}}\big(E_1(\mathbf{I}), E_2(\mathbf{D}), E_3(\mathbf{d}), E_4(\mathbf{m})\big)
    \label{eq:worldlatent_enc}
\end{equation}
and $\mathbf{z}_{\text{4d}} \in \mathbb{R}^{T' \times H' \times W' \times C_{fused}}$ with same spatial and temporal resolution as modality-specific latents and increased fused latent channels $C_{fused}$.

\smallskip
\noindent\textbf{Multi-Modal Decoding and Training Objective.}
To ensure that $\mathbf{z}_{\text{4d}}$ serves as a complete \textit{spatial language} preserving appearance, geometry, and motion, four modality-specific decoders $\{D_i\}_{i=1}^{4}$ independently reconstruct each modality from the shared latent bottleneck: $\hat{\mathbf{X}}_i = D_i(\mathbf{z}_{\text{4d}})$,
yielding reconstructed outputs $(\hat{\mathbf{I}}, \hat{\mathbf{D}}, \hat{\mathbf{d}}, \hat{\mathbf{m}})$. We optimize WorldVAE end-to-end using a composite objective comprising modality reconstruction $\mathcal{L}_1$ loss, perceptual loss (LPIPS \cite{zhang2018unreasonable}), geometric gradient loss, a 3D pointmap loss \cite{lin2025depth} over the decoded depth and ray maps to enforce cross-modal geometric validity, and a KL regularization term:
\begin{equation}
\begin{aligned}
    \mathcal{L}_{\text{VAE}} &= \sum_{i=1}^{4} \lambda_i \|\mathbf{X}_i - \hat{\mathbf{X}}_i\|_1 + \lambda_{\text{p}} \mathcal{L}_{\text{LPIPS}}(\mathbf{I}, \hat{\mathbf{I}}) \\
    &\quad + \lambda_{\text{g}} \mathcal{L}_{\text{grad}}(\mathbf{D}, \hat{\mathbf{D}}) + \lambda_{\text{pts}} \mathcal{L}_{\text{pts}}(\mathbf{P}, \hat{\mathbf{P}}) + \lambda_{\text{KL}} D_{\text{KL}}
\end{aligned}
\label{eq:vae_loss}
\end{equation}
where $\mathcal{L}_{\text{pts}}(\mathbf{P}, \hat{\mathbf{P}}) = \|\mathbf{P} - \hat{\mathbf{P}}\|_1$ supervises the 3D point map $\hat{\mathbf{P}}$ unprojected from the decoded depth $\hat{\mathbf{D}}$ and Pl\"ucker ray maps $(\hat{\mathbf{d}}, \hat{\mathbf{m}})$, $D_{\text{KL}}$ term regularizes the fused latent $z_{4d}$ to match a standard normal distribution $\mathcal{N}(\mathbf{0}, \mathbf{I})$.

\smallskip
\noindent\textbf{Partial \methodName{} Encoding and Decoding.}
Leveraging the flexibility of our DPT token fusion mechanism, we apply random modality-level token masking during training, where unprovided modality streams are replaced with learnable mask tokens prior to fusion. This equips WorldVAE with the ability to encode and decode arbitrary subsets of input modalities into a partial \methodName{} representation $\mathbf{c}_{4d}$, seamlessly bridging full-modality reconstruction in stage (a) and partial-modality conditioning in stage (b).

\subsection{Unified Any-to-4D Generative Modeling}
\label{sec:stage2}

With the pre-trained WorldVAE frozen and \methodName{} serving as the shared representation protocol, Stage (b) (Figure \ref{fig:pipeline} (b)) trains a single generative diffusion model directly within the $\mathbf{z}_{4d}$ latent space to unify 4D perception, generation, and editing in one shared optimization loop.

\smallskip
\noindent\textbf{Shared Representation for Multi-Task Perception and Generation.}
As shown in Figure~\ref{fig:pipeline} (b), both the conditioning input space and the target output space of our unified diffusion model reside entirely within the unified 4D latent space across all 4D perception and generation tasks. Consequently, regardless of the downstream task, the optimization target of the diffusion model remains identical: synthesizing the complete \methodName{} representation $\mathbf{z}_{4d}$, which encodes the full 4D world state.

\smallskip
\noindent\textbf{Partial \methodName{} Conditioning Protocol.}
To enable a single model to accept arbitrary modality subsets, we construct a partial \methodName{} condition $\mathbf{c}_{4d}(\mathcal{S}) \in \mathbb{R}^{T' \times H' \times W' \times C_{fused}}$ by replacing unobserved modalities ($i \notin \mathcal{S}$) with a zero-masked token $\mathbf{0}_i$:
\begin{equation}
    \mathbf{c}_{4d}(\mathcal{S}) = \mathcal{F}_{\text{DPT}}\big(\{\mathbb{I}[i \in \mathcal{S}] \cdot E_i(\mathbf{X}_i) + \mathbb{I}[i \notin \mathcal{S}] \cdot \mathbf{0}_i\}_{i=1}^{4}\big).
    \label{eq:partial_cond}
\end{equation}
As depicted in Figure \ref{fig:pipeline} (b), this unified conditioning interface supports diverse 4D tasks without architectural modifications in a single multi-task training loop:
\begin{itemize}
    \item \textit{4D World Perception ($\mathcal{S} = \{1\}$, RGB-Only)}: Conditioned on monocular video $\mathbf{c}_{4d}(\{1\})$ with $\mathbf{c}_{\text{text}} = \varnothing$, the model predicts the full $\mathbf{z}_{4d}$, from which $D_2, D_3, D_4$ decode 4D depth and 3D camera trajectories.
    \item \textit{Text-to-4D Generation ($\mathcal{S} = \varnothing$, Empty Modality):} Conditioned on a text prompt $\mathbf{c}_{\text{text}}$ and $\mathbf{c}_{4d}(\varnothing)$, the model synthesizes $\mathbf{z}_{4d}$ from scratch, jointly decoding RGB video, depth, and camera motion via $D_1$--$D_4$.
    \item \textit{Controllable 4D Editing ($\mathcal{S} = \{2\}$, Depth + Text):} Conditioned on 4D depth latents $\mathbf{c}_{4d}(\{2\})$ and text $\mathbf{c}_{\text{text}}$, the model predicts $\mathbf{z}_{4d}$ to decode geometrically consistent, appearance-edited RGB videos via $D_1$.
\end{itemize}

\smallskip
\noindent\textbf{Single-Loop Multi-Task Diffusion Optimization.}
Given a clean full latent $\mathbf{z}_{4d}$ and diffusion timestep $\tau \in [0, 1]$, we form the noisy latent $\mathbf{z}_\tau = \alpha_\tau \mathbf{z}_{4d} + \sigma_\tau \boldsymbol{\epsilon}$ with $\boldsymbol{\epsilon} \sim \mathcal{N}(\mathbf{0}, \mathbf{I})$. We concatenate $\mathbf{z}_\tau$ and $\mathbf{c}_{4d}(\mathcal{S})$ along the channel dimension as input to the diffusion backbone $\mathbf{v}_\theta$, while injecting text embeddings $\mathbf{c}_{\text{text}}$ via cross-attention. Across each training batch, we sample tasks $(\mathcal{S}, \mathbf{c}_{\text{txt}}) \sim p_{\text{task}}$ and minimize a single unified denoising loss:
\begin{equation}
    \mathcal{L}_{\text{diff}}(\theta) = \mathbb{E} \Big[ \big\| \mathbf{v}_\theta\big(\mathbf{z}_\tau , \mathbf{c}_{\text{4d}}(\mathcal{S}), \tau, \mathbf{c}_{\text{txt}}\big) - \mathbf{v}_{\text{target}} \big\|_2^2 \Big]
    \label{eq:diff_loss}
\end{equation}
where $\tau \in [0, 1]$, and $(\mathcal{S}, \mathbf{c}_{\text{txt}}) \sim p_{\text{task}}$ task distribution, $\mathbf{v}_\theta(\cdot)$ and $\mathbf{v}_{\text{target}}$ are predicted and ground-truth noise or velocity.
At inference time, iterative denoising from Gaussian noise $\mathbf{z}_T \sim \mathcal{N}(\mathbf{0}, \mathbf{I})$ produces ${\mathbf{z}}_{4d}$, which is decoded by the frozen decoders $D_1$--$D_4$ into synchronized video, dense depth, and camera trajectories.

%% file: sec/4_experiments.tex
\section{Experiments}
\label{sec:experiments}

The results and comparisons are best viewed through video examples. We include video results in our project webpage.

\subsection{Experimental Setup}
\label{sec:exp_setup}

\noindent\textbf{Implementation Details.}
WorldVAE initializes all four encoder-decoder pairs from the Wan VAE \cite{wan2025wan}, freezing the encoders while fine-tuning the decoders alongside a 20M-parameter DPT Fusion module. For \textit{Any-to-4D}, we fine-tune all parameters of Wan2.1-T2V-14B to generate 17-frame, $512{\times}512$ sequences across RGB, depth, and rays.

\smallskip
\noindent\textbf{Datasets.}
We train on $\sim$1.4M synthetic and real-world 17-frame clips, using DA3-Giant~\cite{lin2025depth} to label depth and camera poses on clips lacking dense 4D ground truth. Full training and dataset details in the supplementary material.

\subsection{4D WorldVAE Reconstruction Quality}
\label{sec:exp_vae}

We first evaluate the representational capacity and computational efficiency of WorldVAE in Table~\ref{tab:vae_roundtrip}. A naive approach to joint 4D generation is to tokenize RGB, depth, and camera rays as separate latent streams, which triples the sequence length ($15{,}360$ vs.\ $5{,}120$ DiT tokens) and incurs a $3.84\times$ computational overhead ($569.1$ vs.\ $148.1$ TFLOPs per NFE on a 14B DiT) due to quadratic self-attention scaling. In contrast, WorldVAE fuses all three modalities into a single $5{\times}64{\times}64{\times}32$ latent grid, preserving the exact $5{,}120$-token count of the RGB-only VAE. Because widening the latent channels from $16$ to $32$ only affects the initial patch-embedding and final projection layers of the DiT, WorldVAE adds a negligible $+0.0165\%$ TFLOPs overhead.

Despite compressing four heterogeneous 4D modalities into a single compact bottleneck, WorldVAE achieves strong zero-shot round-trip reconstruction across all domains on unseen evaluation benchmarks, including MPI Sintel and the ScanNet test split: it retains high visual fidelity ($32.23$~dB PSNR, $0.888$ SSIM) alongside accurate metric depth recovery ($0.056$ AbsRel, $0.984$ $\delta_1$) and precise camera ray reconstruction ($1.46^\circ$ ray angular error, $1.77^\circ$ relative rotation error). This confirms that $\mathbf{z}_{\text{4d}}$ serves as a compact, loss-tolerant spatial language that generalizes well for downstream Any-to-4D diffusion modeling.

\begin{table}[t]
\centering

\footnotesize
\setlength{\tabcolsep}{3.5pt}
\renewcommand{\arraystretch}{1.12}

\resizebox{\columnwidth}{!}{%
\begin{tabular}{lccc}
\toprule
\textbf{Tokenizer} & \textbf{Latent Shape} & \textbf{DiT Tokens} & \textbf{TFLOPs / NFE (14B)} \\
\midrule
Wan2.1 RGB VAE (ref.) & $5{\times}64{\times}64{\times}16$ & $5{,}120$ & $148.1$ ($1.00\times$) \\
Separate RGB+D+Ray & $3{\times}(5{\times}64{\times}64{\times}16)$ & $15{,}360$ & $569.1$ ($3.84\times$) \\
\midrule
\textbf{WorldVAE (Ours)} & $5{\times}64{\times}64{\times}32$ & $5{,}120$ & $148.1$ ($+0.02\%$) \\
\bottomrule
\end{tabular}%
}

\vspace{3pt}

\resizebox{\columnwidth}{!}{%
\begin{tabular}{lccccccc}
\toprule
& \multicolumn{3}{c}{\textbf{RGB Video}} & \multicolumn{2}{c}{\textbf{4D Depth}} & \multicolumn{2}{c}{\textbf{Camera Rays}} \\
\cmidrule(lr){2-4} \cmidrule(lr){5-6} \cmidrule(lr){7-8}
\textbf{Tokenizer} & PSNR$\uparrow$ & SSIM$\uparrow$ & LPIPS$\downarrow$ & AbsRel$\downarrow$ & $\delta_1\uparrow$ & Ray$^\circ\downarrow$ & RRE$^\circ\downarrow$ \\
\midrule
Wan2.1 RGB VAE (ref.) & 35.13 & 0.908 & 0.126 & --- & --- & --- & --- \\
\midrule
\textbf{WorldVAE (Ours)} & 32.23 & 0.888 & 0.184 & 0.056 & 0.984 & 1.46 & 1.77 \\
\bottomrule
\end{tabular}%
}
\caption{{\bf 4D WorldVAE reconstruction quality and efficiency}. WorldVAE jointly compresses RGB, depth, and camera rays into the exact the numbers of DiT tokens as an RGB-only VAE while enabling high-fidelity 4D reconstruction.}
\label{tab:vae_roundtrip}
\end{table}

\subsection{Any-to-4D: 4D Perception}
\label{sec:exp_perception}

To evaluate how well our unified representation protocol supports physical scene understanding, we benchmark \methodName{}-14B \emph{zero-shot} across three dynamic and static 4D benchmarks: MPI Sintel \cite{butler2012naturalistic}, Bonn RGB-D \cite{palazzolo2019refusion}, and ScanNet \cite{dai2017scannet}. Table~\ref{tab:perception_benchmarks} compares our 4D generalist model against feed-forward 3D/4D geometry and pose specialists and diffusion depth specialists.

\begin{figure*}[t]
\centering
\includegraphics[width=\textwidth]{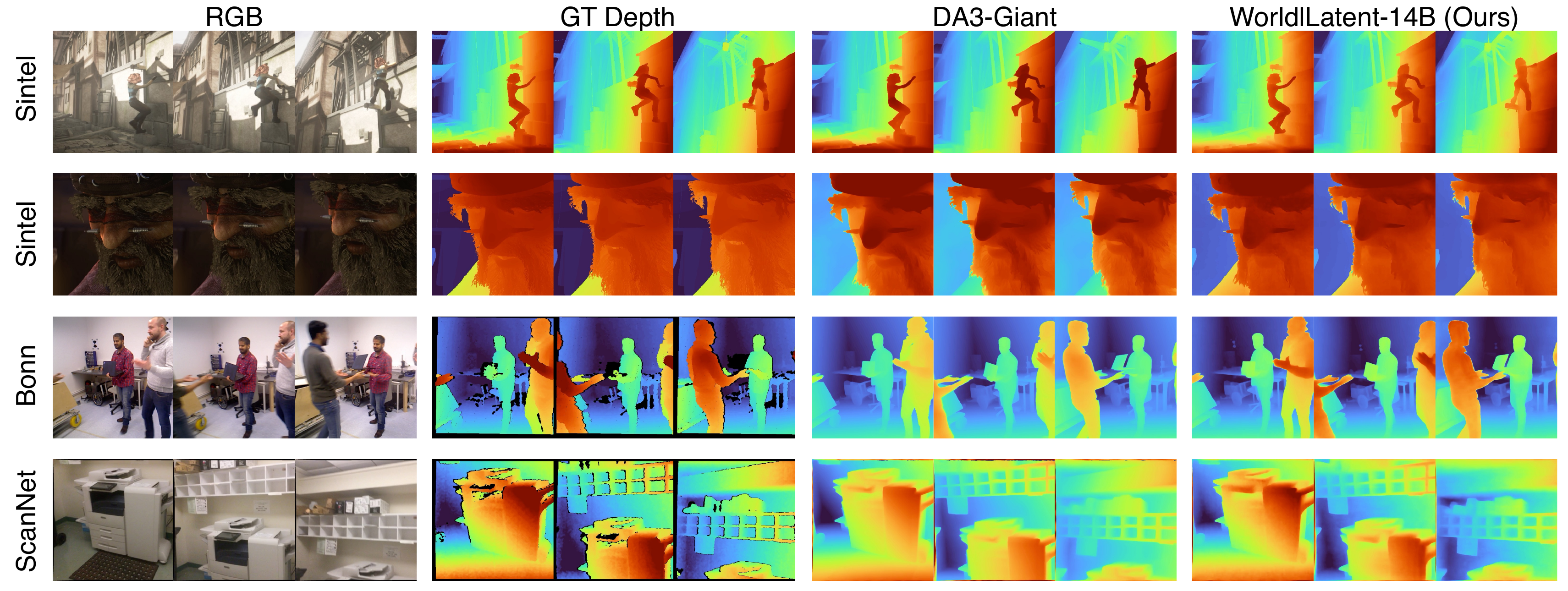}
\caption{\textbf{Qualitative comparison of zero-shot video depth estimation.} Results are shown across the MPI Sintel, Bonn, and ScanNet datasets. For each clip, we display the input RGB frames, ground-truth depth, predictions from the teacher model (DA3-Giant) and \methodName{}-14B (Ours). }
\label{fig:exp_perception}
\end{figure*}

\smallskip
\noindent\textbf{Competitive Performance Against 3D/4D Specialists.}
Operating in just $8$ NFE (number of denoising steps) without global alignment, on Sintel and ScanNet, \methodName{}-14B outperforms diffusion baselines (DepthCrafter and Marigold-v1.1) and several established 3D/4D reconstruction specialists (DUSt3R, MASt3R, and MonST3R) in both depth and camera pose estimation. Notably, on the challenging dynamic MPI Sintel benchmark, \methodName{}-14B achieves the best $\delta_1$ depth accuracy ($0.769$) across all evaluated methods, surpassing even its feed-forward teacher DA3-Giant ($0.763$), Video-Depth-Anything-L ($0.736$), and VGGT-1B ($0.705$), while securing the second-best camera pose $\text{AUC@30}$ ($0.701$) and outperforming all video depth specialists in $\text{AbsRel}$ ($0.247$). On ScanNet, our model ties DA3-Giant for the second-best $\text{AbsRel}$ ($0.042$) with $0.978$ $\delta_1$, and on Bonn it achieves the second-best camera pose $\text{AUC@30}$ ($0.918$) alongside $0.067$ $\text{AbsRel}$ and $0.961$ $\delta_1$. As a generalist 4D world model, \methodName{} performs comparably to state-of-the-art 4D perception specialists while uniquely supporting text-to-4D generation and controllable 4D re-rendering ($\text{D} / \text{P} / \text{G} / \text{R}$) from the same unified model. Figure~\ref{fig:exp_perception} presents qualitative comparisons between our \methodName{}-14B and DA3-Giant across multiple datasets.

\begin{table*}[t]
\centering
\setlength{\tabcolsep}{6pt}
\renewcommand{\arraystretch}{1.15}
\resizebox{\textwidth}{!}{%
\begin{tabular}{l c c ccc}
\toprule
\textbf{Method}
& \textbf{Capabilities}
& \textbf{NFE}
& \textbf{Sintel (Dynamic)}
& \textbf{Bonn (Dynamic)}
& \textbf{ScanNet (Indoor)} \\
& \small{(D / P / G / R)}
& \small{Steps}
& \small{$\text{AbsRel} \downarrow$ / $\delta_1 \uparrow$ / $\text{RRE}^\circ \downarrow$ / $\text{AUC@30} \uparrow$}
& \small{$\text{AbsRel} \downarrow$ / $\delta_1 \uparrow$ / $\text{RRE}^\circ \downarrow$ / $\text{AUC@30} \uparrow$}
& \small{$\text{AbsRel} \downarrow$ / $\delta_1 \uparrow$ / $\text{RRE}^\circ \downarrow$ / $\text{AUC@30} \uparrow$} \\
\midrule
\midrule
\multicolumn{6}{l}{\textit{Feed-Forward 3D/4D Perception Specialists}} \\
\midrule
\midrule
DUSt3R \cite{wang2024dust3r}
& \checkmark\ / \checkmark\ / $\times$\ / $\times$
& 1 (+GA)
& $0.343$ / $0.515$ / $15.60$ / $0.276$
& $0.163$ / $0.748$ / $3.23$ / $0.779$
& $0.050$ / $0.975$ / $2.79$ / $0.445$ \\
MASt3R \cite{leroy2024grounding}
& \checkmark\ / \checkmark\ / $\times$\ / $\times$
& 1 (+GA)
& $0.314$ / $0.566$ / $10.78$ / $0.295$
& $0.193$ / $0.690$ / $2.63$ / $0.802$
& $0.056$ / $0.966$ / $3.05$ / $0.410$ \\
MonST3R \cite{zhang2025monst3r}
& \checkmark\ / \checkmark\ / $\times$\ / $\times$
& 1 (+GA)
& $0.297$ / $0.565$ / $8.10$ / $0.368$
& $0.113$ / $0.854$ / $1.55$ / $0.823$
& $0.065$ / $0.953$ / $3.11$ / $0.408$ \\
CUT3R-512 \cite{wang2025continuous}
& \checkmark\ / \checkmark\ / $\times$\ / $\times$
& 1
& $0.356$ / $0.569$ / $3.61$ / $0.505$
& $\underline{0.059}$ / $0.969$ / $0.84$ / $0.858$
& $0.044$ / $\underline{0.982}$ / $1.65$ / $0.624$ \\
VGGT-1B \cite{wang2025vggt}
& \checkmark\ / \checkmark\ / $\times$\ / $\times$
& 1
& $\underline{0.216}$ / $0.705$ / $\underline{2.65}$ / $0.645$
& $0.067$ / $0.956$ / $\underline{0.38}$ / $0.914$
& $\mathbf{0.028}$ / $\mathbf{0.988}$ / $\mathbf{1.07}$ / $\mathbf{0.724}$ \\
Video-Depth-Anything-L \cite{chen2025video}
& \checkmark\ / $\times$\ / $\times$\ / $\times$
& 1
& $0.250$ / $0.736$ / --- / ---
& $\mathbf{0.056}$ / $\mathbf{0.973}$ / --- / ---
& $0.046$ / $\underline{0.982}$ / --- / --- \\
Metric-VDA-Large \cite{chen2025video}
& \checkmark\ / $\times$\ / $\times$\ / $\times$
& 1
& $0.264$ / $0.697$ / --- / ---
& $0.071$ / $0.963$ / --- / ---
& $0.052$ / $0.978$ / --- / --- \\
DA3-Giant (Teacher) \cite{lin2025depth}
& \checkmark\ / \checkmark\ / $\times$\ / $\times$
& 1
& $\mathbf{0.194}$ / $\underline{0.763}$ / $\mathbf{1.91}$ / $\mathbf{0.704}$
& $\underline{0.059}$ / $\underline{0.972}$ / $\mathbf{0.36}$ / $\mathbf{0.923}$
& $\underline{0.042}$ / $0.977$ / $\underline{1.12}$ / $\underline{0.703}$ \\
\midrule
\midrule
\multicolumn{6}{l}{\textit{Diffusion Video/Image Depth Specialists}} \\
\midrule
\midrule
Marigold-v1.1 \cite{ke2024repurposing}
& \checkmark\ / $\times$\ / $\times$\ / $\times$
& 50
& $0.392$ / $0.409$ / --- / ---
& $0.281$ / $0.442$ / --- / ---
& $0.191$ / $0.670$ / --- / --- \\
DepthCrafter \cite{hu2025depthcrafter}
& \checkmark\ / $\times$\ / $\times$\ / $\times$
& 25
& $0.294$ / $0.725$ / --- / ---
& $0.066$ / $0.966$ / --- / ---
& $0.081$ / $0.941$ / --- / --- \\
\midrule
\midrule
\multicolumn{6}{l}{\textit{Unified 4D World Model}} \\
\midrule
\midrule
\textbf{\methodName{}-14B (Ours)}
& \checkmark\ / \checkmark\ / \checkmark\ / \checkmark
& 8
& $0.247$ / $\mathbf{0.769}$ / $3.12$ / $\underline{0.701}$
& $0.067$ / $0.961$ / $0.60$ / $\underline{0.918}$
& $\underline{0.042}$ / $0.978$ / $2.66$ / $0.640$ \\
\bottomrule
\end{tabular}%
}
\caption{\textbf{Comparison with state-of-the-art methods on 4D perception.} Zero-shot evaluation across dynamic and static multi-domain datasets under scale-and-shift alignment. Evaluated capabilities include (1) \textbf{D}epth estimation, (2) camera \textbf{P}ose estimation, (3) 4D \textbf{G}eneration, and (4) controllable \textbf{R}e-rendering/editing from depth maps. \texttt{+GA} denotes the test-time global alignment optimization used by baseline methods. \textbf{Bold} and \underline{underlined} values indicate the best and second-best results, respectively. \methodName{} performs comparably to dedicated 3D/4D perception specialists, while providing additional capabilities for generation and re-rendering.}
\label{tab:perception_benchmarks}
\end{table*}

\subsection{Any-to-4D: 4D World Generation and Editing}
\begin{figure*}[t]
\centering
\includegraphics[width=0.95\textwidth]{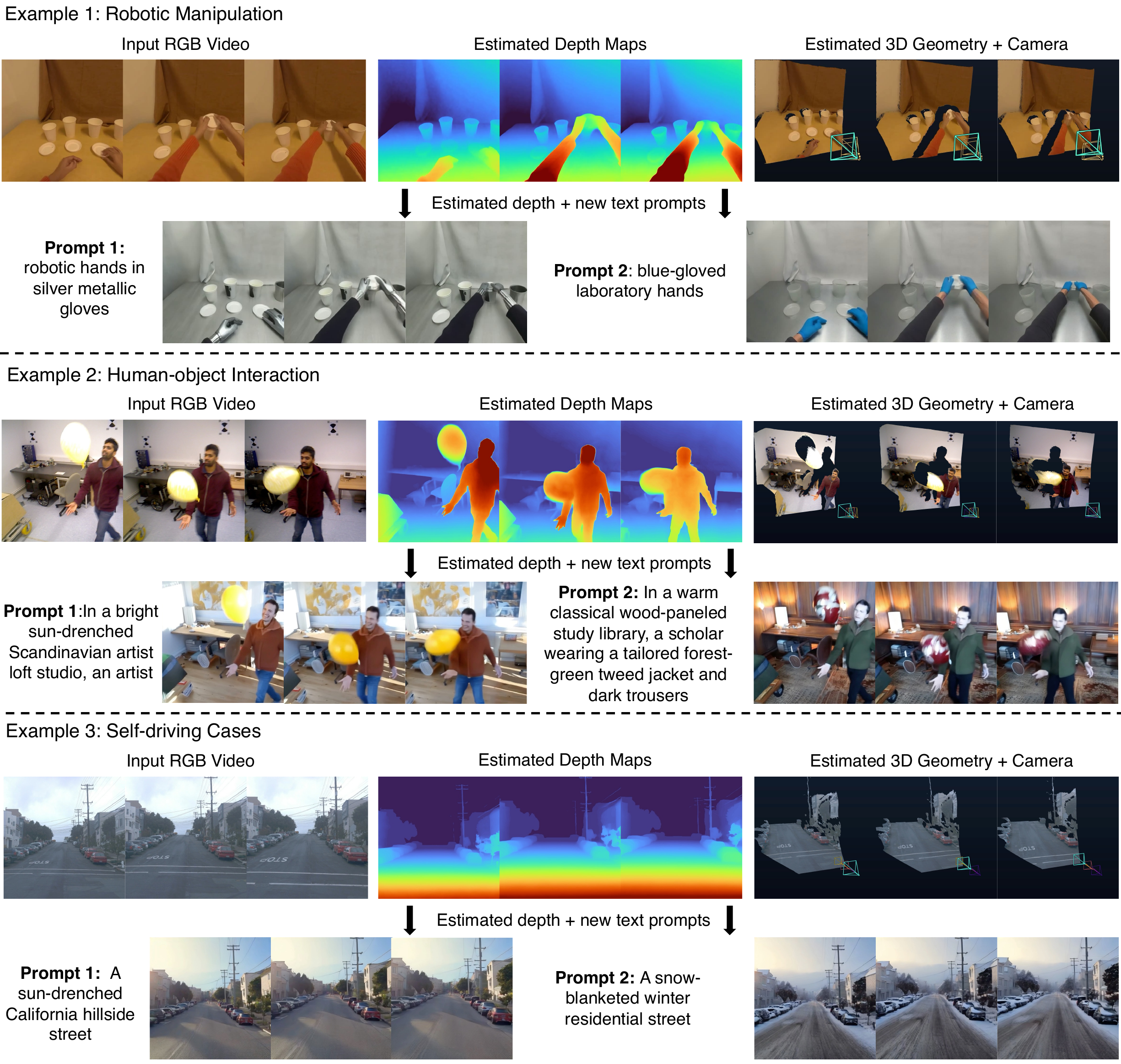}
\caption{\textbf{Any-to-4D: world editing and depth-conditioned re-rendering with \methodName{}.} (1) Given an input RGB video, our model first performs 4D perception to estimate dense depth maps. (2) Conditioned on these predicted depth maps and new text prompts, the same model then performs 4D re-rendering to synthesize appearance-edited RGB videos. For each example, we show the input video, the predicted depth and camera outputs from our 4D perception stage, and the re-rendered videos across diverse text prompts.}
\label{fig:exp_render}

\end{figure*}

\begin{figure*}[t]
\centering
\includegraphics[width=0.9\textwidth]{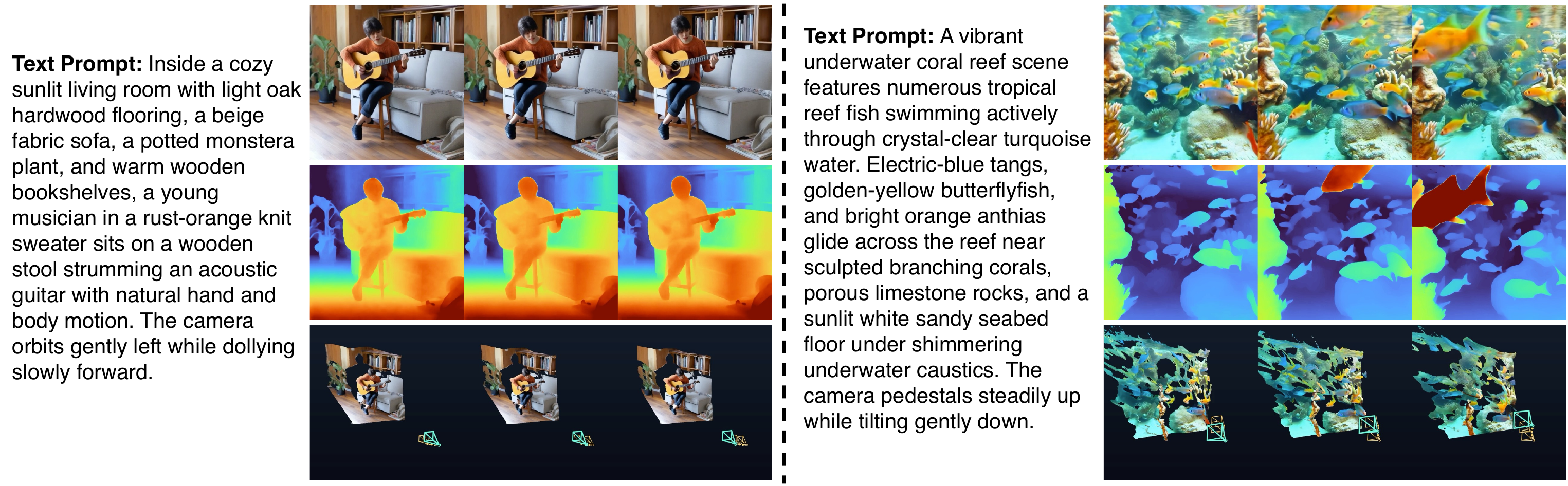}
\caption{{\bf Any-to-4D: text-to-4D world generation with \methodName{}}. Given only a text prompt, our model jointly denoises appearance, depth, and camera rays from noise, producing an RGB video together with aligned depth maps and a 3D scene with camera trajectory.}
\label{fig:exp_gen}
\end{figure*}

Using the same \methodName{}-14B model as in Section~\ref{sec:exp_perception}, we evaluate text-to-4D generation (Figure~\ref{fig:exp_gen}) and depth-conditioned 4D editing/re-rendering (Figure~\ref{fig:exp_render}). To showcase the unification of perception and generation within a single model, our 4D editing results execute a self-contained chained pipeline: (1) given an input RGB video, \methodName{}-14B first performs 4D perception to infer depth maps, and (2) conditioned on these predicted depth maps and new text prompts, the same model re-renders geometrically consistent, appearance-edited 4D videos.

Table~\ref{tab:generation_benchmarks} quantifies the benefits of our unified 4D latent space over the RGB-only model across $55$ held-out video-prompt pairs using VBench \cite{huang2024vbench} and 3D consistency metrics. While the base model attains superficially high smoothness and flicker scores by generating nearly static cameras ($0.058$ path, $0.99^\circ$ rotation) and low motion ($0.400$ Dyn), \methodName{}-14B synthesizes rich 3D camera trajectories ($0.401 / 30.45^\circ$ for generation; $0.445 / 26.06^\circ$ for editing) and high scene dynamics ($0.964$ Dyn) that closely match real-world videos ($0.485 / 24.55^\circ$, $0.945$ Dyn). Crucially, despite a $30\times$ increase in camera rotation, our model maintains real-video-level 3D geometric consistency ($\sim\!0.023$ warp / $0.011$ drift vs.\ $0.022 / 0.013$) and surpasses the base model in background consistency ($0.953$ vs.\ $0.944$).

\begin{table}[t]
\centering
\setlength{\tabcolsep}{2.5pt}
\renewcommand{\arraystretch}{1.15}
\resizebox{\columnwidth}{!}{%
\begin{tabular}{l c c c c c}
\toprule
\textbf{Method}
& \textbf{Cond.}
& \textbf{Camera Motion}
& \textbf{CLIP-T} $\uparrow$
& \textbf{VBench Quality \& Dynamics} $\uparrow$
& \textbf{3D Consistency} $\downarrow$ \\
& \small{Input}
& \small{Path / Rot$^\circ$}
& \small{Score}
& \small{Subj / Bg / Flick / Smooth / Dyn}
& \small{Warp / Drift} \\
\midrule
Wan2.1-T2V-14B (Base)
& Text
& $0.058$ / $0.99^\circ$
& $0.264$
& $.911$ / $.944$ / $.979$ / $.990$ / $.400$
& $.0163$ / $.0053$ \\
\midrule
\textbf{\methodName{}-14B (Gen.)}
& Text
& $0.401$ / $30.45^\circ$
& $0.243$
& $.856$ / $.953$ / $.942$ / $.970$ / $.964$
& $.0228$ / $.0112$ \\
\textbf{\methodName{}-14B (Edit)}
& Depth+Text
& $0.445$ / $26.06^\circ$
& $0.235$
& $.831$ / $.950$ / $.943$ / $.967$ / $.964$
& $.0233$ / $.0124$ \\
\midrule
\textit{Real Videos (Reference)}
& ---
& $0.485$ / $24.55^\circ$
& $0.253$
& $.873$ / $.952$ / $.928$ / $.958$ / $.945$
& $.0220$ / $.0130$ \\
\bottomrule
\end{tabular}%
}
\caption{\textbf{Quantitative evaluation of Any-to-4D generation and depth-controllable editing.} Unlike the base Wan model, which produces nearly static cameras ($0.99^\circ$ rotation, $0.400$ dynamic degree), \methodName{}-14B synthesizes rich camera trajectories and scene dynamics that match real-world videos while preserving strong 3D consistency and background coherence.}
\label{tab:generation_benchmarks}
\end{table}

\subsection{Ablation Studies}
\label{sec:ablations}

\begin{figure}[t]
\centering
\includegraphics[width=\columnwidth]{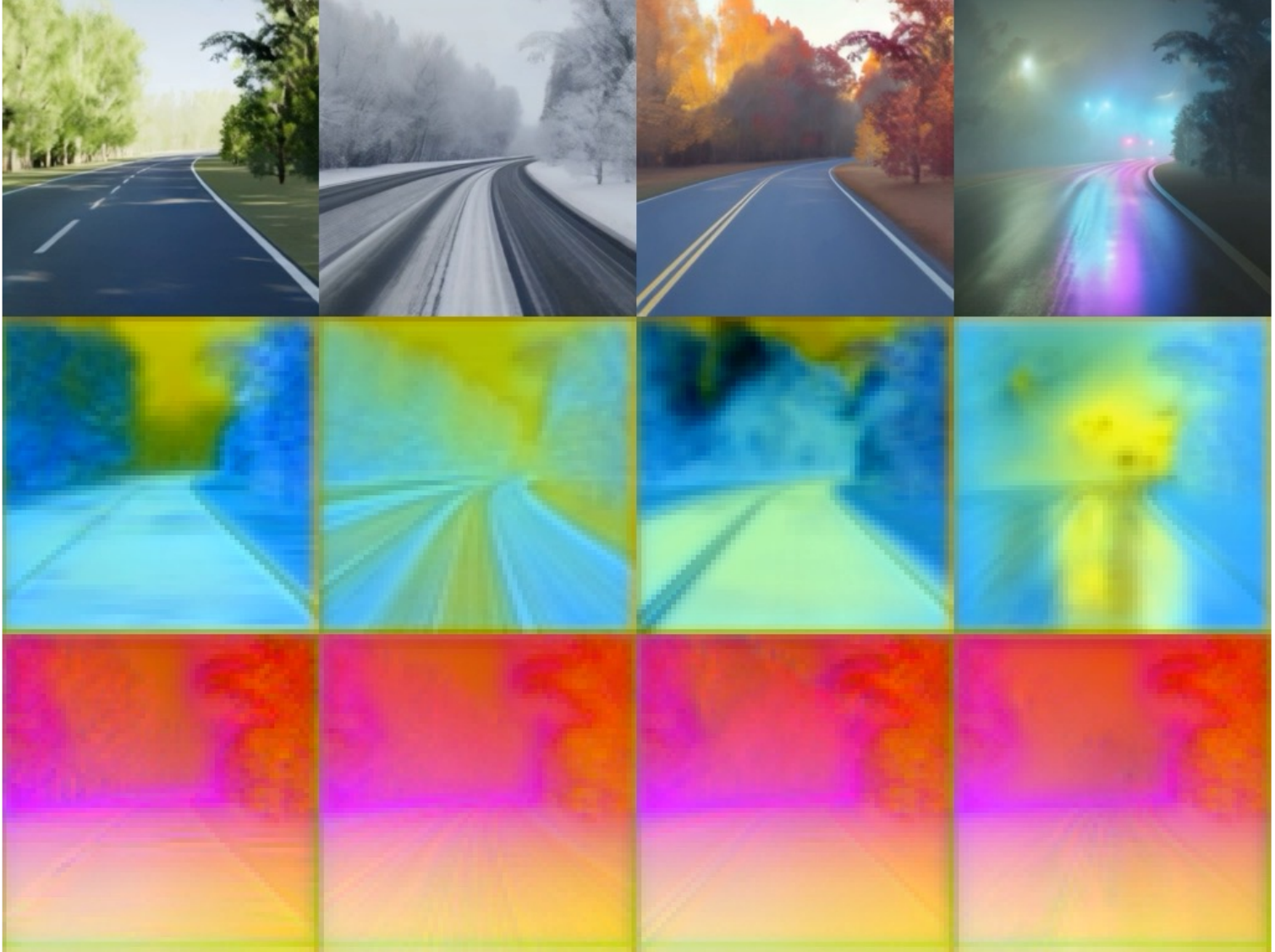}
\caption{\textbf{Linear subspace decomposition of \methodName{}.} Visualization across videos re-rendered from an identical depth sequence under different text prompts (\textit{Row 1}). The $16$-D appearance projection $\Pi_\parallel \mathbf{z}_{\text{4d}}$ (\textit{Row 2}) varies markedly with visual style, whereas the orthogonal $16$-D geometry-and-camera projection $\Pi_\perp \mathbf{z}_{\text{4d}}$ (\textit{Row 3}) remains nearly invariant.}
\label{fig:abla_decomp}
\end{figure}

\noindent\textbf{Linear Subspace Decomposition of \methodName{}.}
Fitting a closed-form ridge projection $\mathbf{W} \in \mathbb{R}^{32 \times 16}$ from our $32$-ch $\mathbf{z}_{\text{4d}}$ to the pretrained $16$-ch Wan2.1 RGB latent $\mathbf{z}_v$ yields $R^2 = 0.96$. Using the rank-$16$ orthogonal projector $\mathbf{P}_W = \mathbf{W}(\mathbf{W}^\top \mathbf{W})^{-1}\mathbf{W}^\top$, each token $(t, h, w)$ decomposes as $\mathbf{z}_{\text{4d}} = \Pi_\parallel \mathbf{z}_{\text{4d}} + \Pi_\perp \mathbf{z}_{\text{4d}}$ into a $16$-D \textit{appearance subspace} ($\Pi_\parallel = \mathbf{P}_W$, aligned with RGB) and an orthogonal $16$-D \textit{geometry-and-camera subspace} ($\Pi_\perp = \mathbf{I} - \mathbf{P}_W$, encoding depth and Pl\"ucker rays). As shown in Figure~\ref{fig:abla_decomp}, re-rendering the same depth trajectory with distinct text prompts induces substantial changes in $\Pi_\parallel \mathbf{z}_{\text{4d}}$ while leaving $\Pi_\perp \mathbf{z}_{\text{4d}}$ nearly unchanged, confirming that \methodName{} learns a cleanly disentangled, geometry-locked 4D representation within a single $32$-ch tensor.

\smallskip
\noindent\textbf{Single-Loop Multi-Task Unification vs.\ Single-Task Specialists.} We conduct controlled ablations using the Wan2.1-T2V-1.3B backbone under an identical training budget to evaluate our multi-task formulation and unified latent bottleneck design (Table~\ref{tab:ablations}). In Table~\ref{tab:ablations}(a), we compare our jointly trained unified model (\textbf{J}) against three separate single-task 1.3B specialists trained exclusively on perception (\textbf{P}), depth-conditioned rendering (\textbf{R}), or text-to-4D generation (\textbf{G}). While each specialist dedicates $100\%$ of its capacity to a single task and cannot execute the other tasks, our unified 1.3B checkpoint simultaneously handles all three task families with competitive performance across every metric (e.g., matching specialist \textbf{G} in text alignment at $.219$ vs.\ $.220$ CLIP-T and $.945$ dynamic degree, and matching specialist \textbf{R} at $.960$ vs.\ $.964$ render $\delta_1$). Moreover, as demonstrated in Tables~\ref{tab:perception_benchmarks} and~\ref{tab:generation_benchmarks}, scaling the unified model from 1.3B to 14B parameters resolves this capacity bottleneck and matches or surpasses state-of-the-art specialists.

\smallskip
\noindent\textbf{Fused 32-Channel \methodName{} vs.\ Naive 64-Channel Concatenation.}
In Table~\ref{tab:ablations} (b), we ablate our DPT Fusion module against a baseline that naively concatenates the four $16$-channel modality latents along the channel dimension into a $64$-channel latent space (\textit{Separate 64-ch}). Despite using half the latent channels ($32$ vs.\ $64$), our Fused 32-ch representation achieves higher depth inlier accuracy ($.777$ vs.\ $.768$ $\delta_1$), lower ray direction error ($5.03^\circ$ vs.\ $5.90^\circ$), lower temporal drift ($.095$ vs.\ $.108$), and a substantial $+9.3$-point improvement in camera pose $\text{AUC@30}$ ($.692$ vs.\ $.599$). This confirms that cross-modal DPT fusion eliminates inter-modality redundancy and enforces geometric-kinematic coupling that makes the latent space significantly easier for the diffusion model to denoise.

\begin{table}[t]
\centering
\setlength{\tabcolsep}{2.5pt}
\renewcommand{\arraystretch}{1.12}

\newsavebox{\ablTabA}\newsavebox{\ablTabB}\newlength{\ablTabW}
\sbox{\ablTabA}{%
\begin{tabular}{l c c c}
\toprule
\multicolumn{4}{l}{\textbf{(a) Multi-Task Unification vs.\ Single-Task Specialists (1.3B)}} \\
\midrule
\textbf{Training Arm}
& \textbf{4D Perception}
& \textbf{4D Re-rendering}
& \textbf{Text-to-4D Generation} \\
& \small{$\text{AbsRel}\downarrow$ / $\delta_1\uparrow$ / $\text{AUC}\uparrow$}
& \small{$\text{AbsRel}\downarrow$ / $\delta_1\uparrow$ / $\text{PSNR}\uparrow$}
& \small{$\text{CLIP}\uparrow$ / $\text{Warp}\downarrow$ / $\text{Drift}\downarrow$ / $\text{Subj}\uparrow$ / $\text{Dyn}\uparrow$} \\
\midrule
P --- Perception only
& $.186$ / $.811$ / $.766$
& ---
& --- \\
R --- Render only
& ---
& $.065$ / $.964$ / $12.29$
& --- \\
G --- Generation only
& ---
& ---
& $.220$ / $.0464$ / $.0198$ / $.782$ / $.945$ \\
\midrule
\textbf{J --- Unified (Ours)}
& $.216$ / $.773$ / $.726$
& $.079$ / $.960$ / $11.99$
& $.219$ / $.0672$ / $.0262$ / $.741$ / $.945$ \\
\bottomrule
\end{tabular}%
}
\setlength{\ablTabW}{\wd\ablTabA}

\sbox{\ablTabB}{%
\begin{tabular*}{\ablTabW}{@{\extracolsep{\fill}} l ccccccc}
\toprule
\multicolumn{8}{l}{\textbf{(b) Unified DPT-Fused Latent vs.\ Separate Concatenated Latents (1.3B)}} \\
\midrule
\textbf{Latent Space}
& $\text{AbsRel}\downarrow$
& $\delta_1\uparrow$
& $\text{Ray}^\circ\downarrow$
& $\text{RRE}^\circ\downarrow$
& $\text{AUC@30}\uparrow$
& \text{3D Pt.\ Err.}$\downarrow$
& \text{Drift}$\downarrow$ \\
\midrule
Separate ($64$-ch)
& $\mathbf{.205}$ & $.768$ & $5.90$ & $\mathbf{4.12}$ & $.599$ & $\mathbf{.419}$ & $.108$ \\
\textbf{Fused $32$-ch (Ours)}
& $.221$ & $\mathbf{.777}$ & $\mathbf{5.03}$ & $4.55$ & $\mathbf{.692}$ & $.427$ & $\mathbf{.095}$ \\
\bottomrule
\end{tabular*}%
}
\ifdim\wd\ablTabB>\ablTabW \setlength{\ablTabW}{\wd\ablTabB}\fi

\resizebox{\dimexpr\columnwidth*\wd\ablTabA/\ablTabW\relax}{!}{\usebox{\ablTabA}}

\vspace{4pt}

\resizebox{\dimexpr\columnwidth*\wd\ablTabB/\ablTabW\relax}{!}{\usebox{\ablTabB}}
\caption{\textbf{Controlled ablation studies} using the \methodName{}-1.3B backbone, evaluating (a) model capacity for multi-task unification, and (b) our unified \methodName{} versus separate latents.}
\label{tab:ablations}
\end{table}

%% file: sec/5_conclusion.tex
\section{Discussion}
\label{sec:discussion}

We introduced \methodName{}, a unified 4D latent representation protocol that bridges world perception and generative modeling. By fusing visual appearance, geometry, and camera dynamics into a single compact \methodName{} bottleneck, our framework establishes a shared \textit{spatial language} across heterogeneous 4D tasks. Building on this, our \textit{Any-to-4D} diffusion formulation consolidates perception, text-to-4D generation, and controllable re-rendering into a single generative model and optimization loop. Extensive experiments validate this unified approach, laying a scalable blueprint toward general-purpose 4D spatial intelligence. While we have validated single-model 4D perception and generation at a moderate scale, the effects of industrial-scale compute, parameters, and data remain unexplored. An open question for future large-scale research is \textit{whether scaling multi-task training over a shared 4D latent space will catalyze emergent spatial and physical reasoning}, analogous to the trajectory from GPT-2~\cite{radford2019language} to modern frontier LLMs.